\documentclass[letterpaper, 10 pt, conference]{ieeeconf}  

\IEEEoverridecommandlockouts                              

\usepackage{graphics} 
\usepackage{epsfig} 
\usepackage{mathptmx} 
\usepackage{times} 
\usepackage{amsmath} 
\usepackage{amssymb}  
\usepackage{caption}
\usepackage{subcaption}
\usepackage{array}
\usepackage{siunitx}
\usepackage[T1]{fontenc}

\title{\LARGE \bf
One Problem, One Solution: Unifying Robot and Environment Design Optimization
}

\author{Jan Baumgärtner$^{1}$, Gajanan Kanagalingam$^{2}$, Alexander Puchta$^{1}$ and Jürgen Fleischer$^{1}$
\thanks{*The authors would like to thank the Ministry of Science, Research and Arts of the Federal State of Baden-Württemberg for the financial support of the project within the InnovationsCampus Future Mobility.
}
\thanks{$^{1}$Jan Baumgärtner, Alexander Puchta and Jürgen Fleischer are with the with wbk Institute of Production Science,
        Karlsruhe Institute of Technology, 76131 Karlsruhe, Germany
        {\tt\small jan.baumgaertner@kit.edu}}%
\thanks{$^{2}$Gajanan Kanagalingam is with the Institute for System Dynamics,
        University of Stuttgart, 70569 Stuttgart, Germany}
}

\begin{document}

\maketitle
\thispagestyle{empty}
\pagestyle{empty}

\begin{abstract}
The task-specific optimization of robotic systems has long been divided into the optimization of the robot and the optimization of the environment.
In this letter, we argue that these two problems are interdependent and should be treated as such.
To this end, we present a unified problem formulation that enables for the simultaneous optimization of both the robot kinematics and the environment.
We demonstrate the effectiveness of our approach by jointly optimizing a robotic milling system.
To compare our approach to the state of the art we also optimize the robot kinematics and environment separately.
The results show that our approach outperforms the state of the art and that simultaneous optimization leads to a much better solution.
\end{abstract}

\section{INTRODUCTION}
Although robots possess the inherent ability to perform a multitude of diverse tasks across various industrial domains,
each individual robot is often limited to the execution of a singular repetitive task.
This has opened the door for specialized robotic systems optimized for specific tasks.
Optimization criteria vary from cycle time~\cite{bachmann} to deformation over the task~\cite{stiffness_placement} and even the mechanical complexity~\cite{co_optimization}.
These include modular systems such as presented in~\cite{modular_robot} but also complete custom solutions as in~\cite{custom_robot}.
On the algorithmic side, this has raised interest in task-specific robot optimization.
Examples include the work by~\cite{co_optimization} and~\cite{whitman_rl} who try to find optimal robot designs for a given task based on modular components.
But in industrial settings, the environment can often also be optimized to improve system performance.
For example by simply moving the task closer to the robot. It seems evident that this would reduce the required robot workspace and thus the required robot size.
This indicates that the optimal robot design is subject to the environment in which it is used.
In the optimal robot design literature, this is often neglected and the robot is assumed to be used in a fixed environment.
Environment optimization meanwhile is itself a well-studied problem in the field of production science.
Here many works have tried to find the optimal placement of workstations relative to a given robot.
This is not trivial since most manufacturing tasks only constrain the robot in 5 degrees of freedom resulting in a redundant system.
Works such as~\cite{stiffness_pose_optimization} have shown that this redundant degree of freedom can have a significant impact on system performance.
Determining the optimal placement thus requires knowing the optimal joint trajectory.
To summarize we have found two optimization problems typically treated as separate in the literature but which are in fact interdependent.
This work aims to unify both problems into a single optimization method that can be applied to different design problems.

Our main contribution is thus a unified problem description presented in section~\ref{sec:problem_formulation}.
This description enables the application of new optimization algorithms explored in section~\ref{sec:example_application}.
We not only use these to optimize continuous robot design spaces but also the kinematic structure of the robot.
Additionally, we demonstrate how the problem can be extended to instead optimize robots built from predefined modules using a simple three-step approach.
Our solutions are benchmarked against the existing state of the art for both placement and robot optimization in section~\ref{sec:experiments}.
We show that our approach outperforms the state of the art in both subdomains and that it can solve the unified problem.
Finally, we discuss the limitations of the problem formulation and our solution, as well as suggest future research directions in section~\ref{sec:limitations}.

\section{RELATED WORK}\label{sec:related_work}
Since this work aims to unify robot and environment optimization it is heavily influenced by both fields.
In this section, we will give a brief overview of the most relevant works in both fields.
\subsection{Robot Design Optimization}
In the field of Robot design optimization, one can differentiate between two main subfields:
Modular robot design tries to build a robot from predefined modules while continuous robot design relies on optimizing continuous kinematic parameters.
Module-based design approaches tend to \textit{grow} additional modules from the base until they reach the target position or trajectory.
This is typically encoded using a reachability metric~\cite{co_optimization} which is either optimized using a heuristic-based search algorithm~\cite{co_optimization} or using reinforcement learning~\cite{whitman_rl}.
Extending these growing approaches to environment optimization is not trivial since the performance of the growing robot is dependent on its optimal placement which will change as it grows, requiring nested optimization loops.
For this reason, we will instead rely on continuous robot design optimization methods.
Continuous robot design mostly focuses on optimizing the kinematics of a robot. These are described using some kind of parameterization the Denavit-Hartenberg parameterization~\cite{task_synthesis} being a popular example because it requires only 4 parameters $(\theta,d,a,\alpha)$ per link, making it very efficient to optimize.
We will also later use this parameterization for our sample problem.
The main challenge in optimizing robot kinematics lies in mapping the desired task space waypoints to the joint and design parameter space of the robot.
Earlier works used inverse kinematics for this purpose by converting the desired task positions to joint space given a set of design parameters~\cite{task_synthesis}.
Since the inverse kinematics has no closed-form solution for most robots gradient-free methods such as particle swarm optimizers~\cite{task_synthesis}~\cite{PSO} had to be used to then optimize the design parameters. These methods are not only very slow, they fail to incorporate that different inverse kinematic solutions might have different performances.

Recent works have thus formulated the problem explicitly using forward kinematics~\cite{whitman} resulting in nonlinear optimization problems (NLP).
These NLP can be solved using gradient-based methods to find not only the optimal design parameters but also the optimal joint trajectory.
Tracking the desired waypoints is achieved using soft constraints forcing the robot end effector near the target waypoints.
This approach was also picked up in~\cite{environment_design} which despite its name only extends it to include tool geometries that consider reaction forces with the environment.
Our approach takes many ideas from these works, such as the use of link length regularization described in~\cite{whitman} but does not use forward kinematics.
Unfortunately, neither approach considers the optimization of the environment around the robot.
\subsection{Environment Optimization}
The field of manufacturing science has a long history of optimizing the environment around a robot stemming from the field of factory planning.
Since the goal generally is to optimize the performance of the cell the focus is on the placement of the functional parts with which the robot interacts.
These will be referred to as workstations in the following.
These workstations are generally abstracted to their respective toolpaths which are known a priori.
This means that formally the problem is to find the optimal placement of toolpaths relative to a given robot.
The optimization thus has to contend with the same mapping problem as in robot design optimization.
In a complete parallel to robot design optimization, the problem was initially formulated using inverse kinematics~\cite{bachmann}.
A noteworthy example is~\cite{bachmann} which optimized a dual arm assembly station. Their goal in this case was to improve the cycle time of the system by minimizing the Euclidean distance between workstations.
But this is only an approximation since the shortest path of the end effector is not necessarily the fastest.
But the euclidean distance is only a rough approximation of the time it takes the robot to reach a position.
Again we would require a trajectory optimization to evaluate a given design, and again this would result in a nested optimization loop.
The solutions commonly used in mobile manipulation base planning suffer from a similar problem.
A great overview of these can be found in~\cite{mobile_manipulation}.
Instead of using forward kinematics to solve this problem like for robot design optimization, recent works have instead used differential kinematics by formulating the placement problem as a modified optimal control problem~\cite{previous_work}.
Our approach is based on a similar differential kinematic formulation but incorporates the placement parameters differently.
\section{PROBLEM FORMULATION}\label{sec:problem_formulation}
\begin{figure*}[ht]
        \centering
        \begin{subfigure}[b]{0.28\textwidth}
                \includegraphics[width=\textwidth]{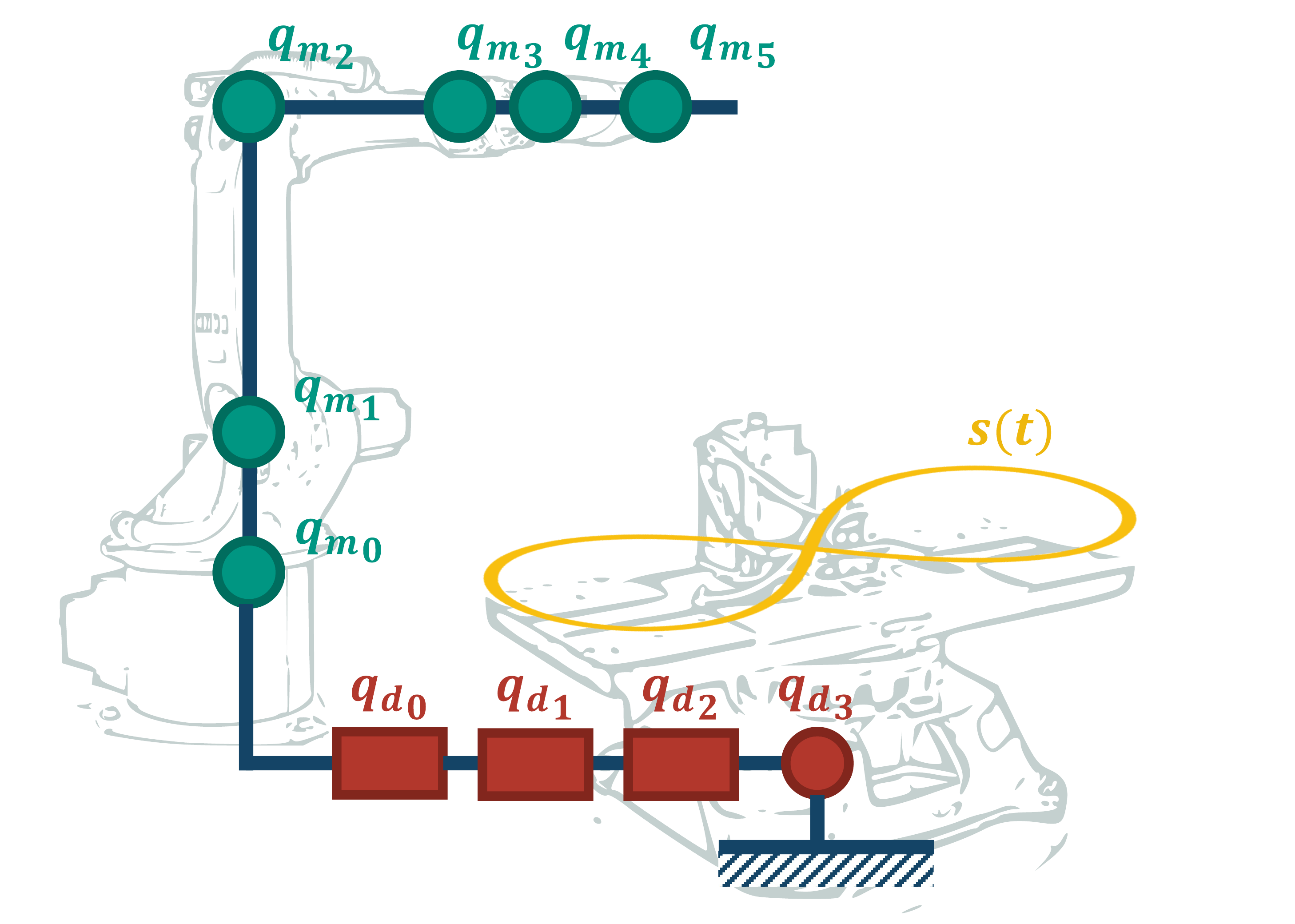}
                \caption{Workpiece placement problem}\label{fig:placement_joints_workpiece}
        \end{subfigure}
        \hfill
        \begin{subfigure}[b]{0.28\textwidth}
                \includegraphics[width=\textwidth]{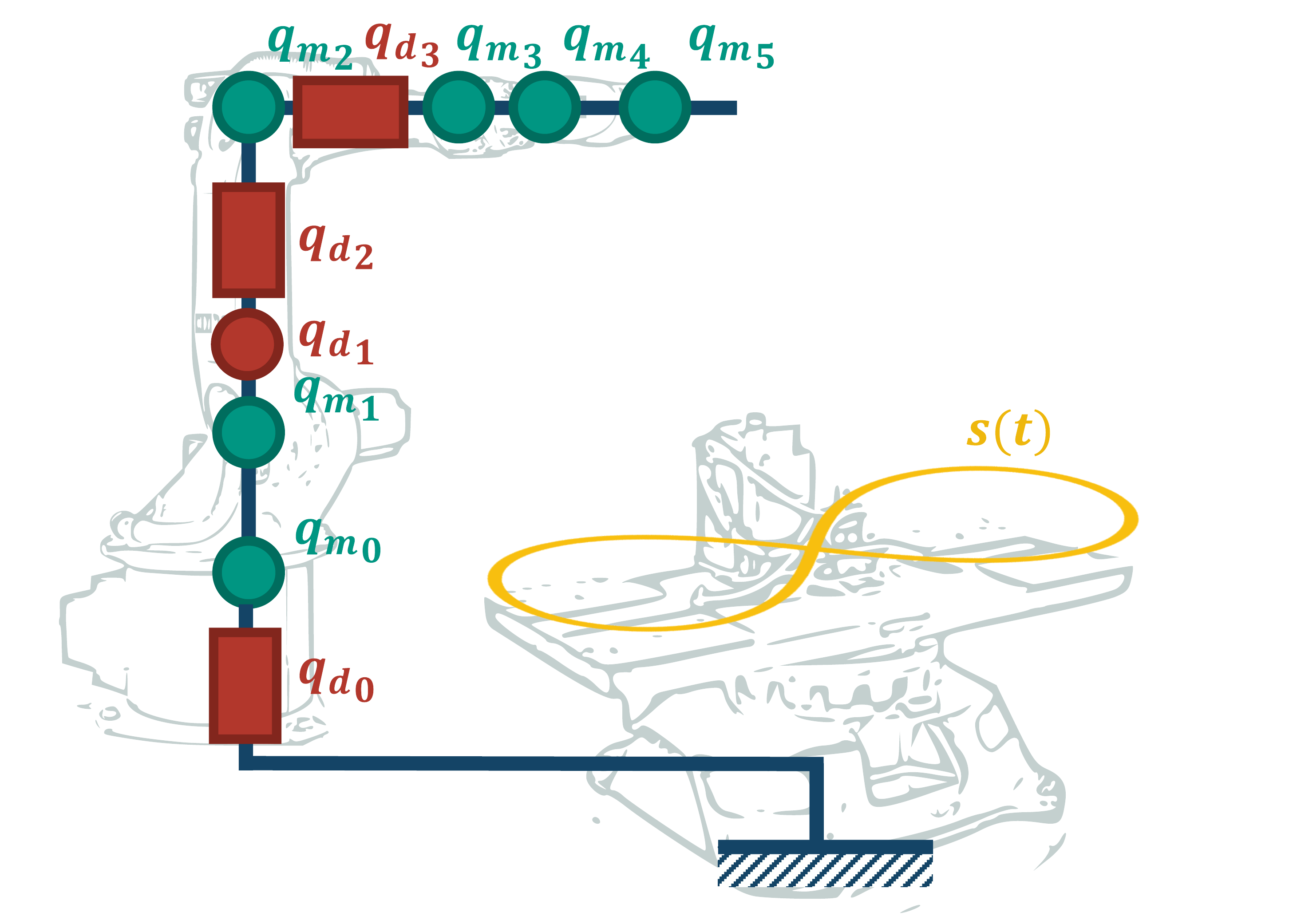}
                \caption{Robot design problem}\label{fig:design_joints_robot}
        \end{subfigure}
        \hfill
        \begin{subfigure}[b]{0.28\textwidth}
                \includegraphics[width=\textwidth]{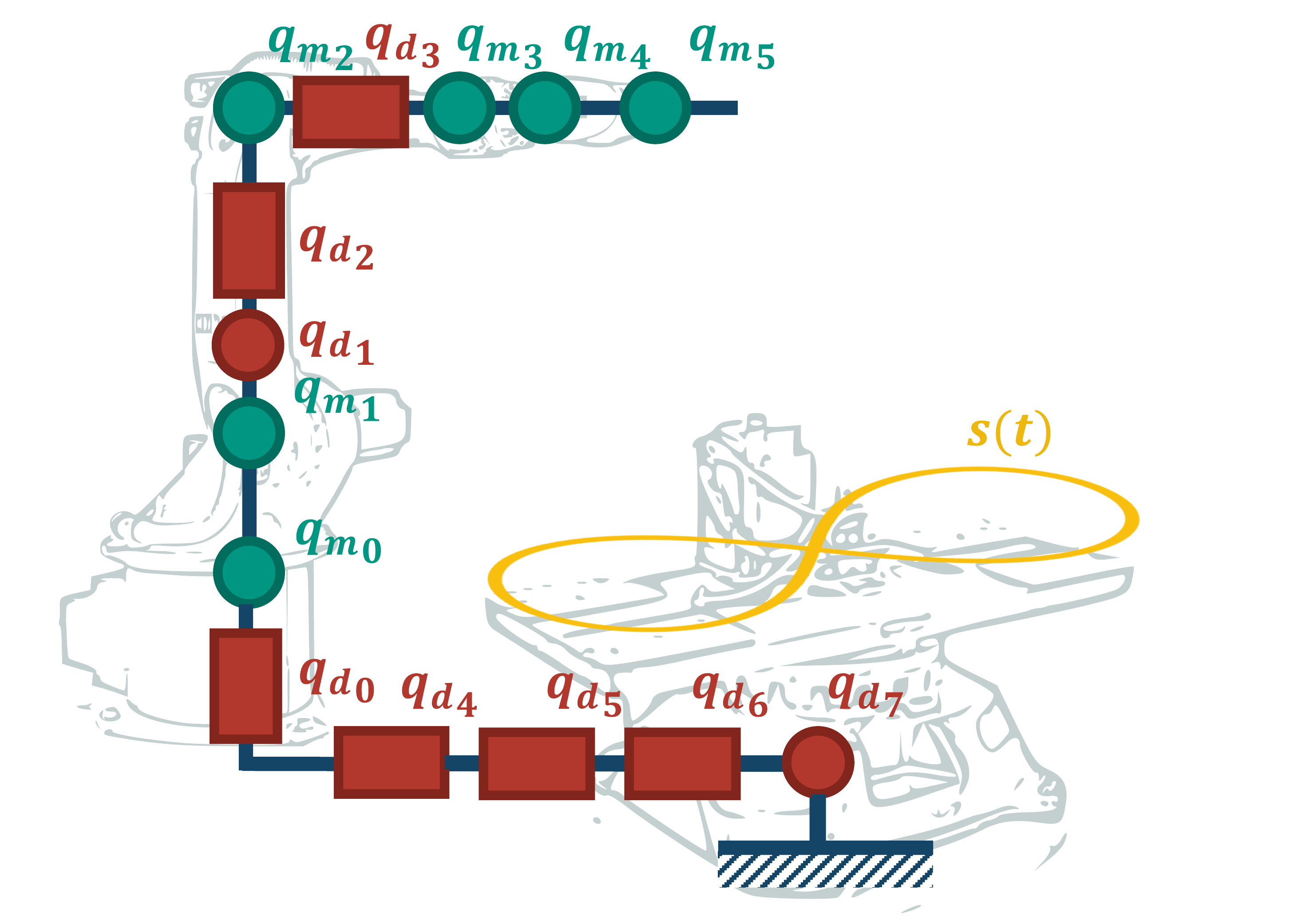}
                \caption{Unified design problem}\label{fig:combined_joints_robot}
        \end{subfigure}
        \caption{Design joints for different design problems}\label{fig:design_joints}

\end{figure*}

Robot design and environment optimization share a remarkably similar history in terms of problem formulation.
Both require finding an optimal joint trajectory to evaluate a given design. In robot design optimization this is done using a forward kinematic formulation~\cite{whitman}.
This results in a very inefficient trajectory optimization because the problem treats subsequent robot positions as uncorrelated.
In general, both robot design and environment optimization have mainly been treated as design problems where trajectory optimization has to be included.
We propose to instead think of these problems as trajectory optimization problems where the design parameters are included as additional joints which we will call design joints.
Link lengths or workpiece positions might be thought of as additional prismatic design joints while the orientation of the workpiece might be thought of as additional revolute design joints.
These design joints are then optimized together with the robot joints with the additional constraint that they have to be constant over time.
Notice that this formulation does not differentiate between robot and environment design parameters, depending on where they are inserted the formulation can be used for both.
Fig.~\ref{fig:design_joints} illustrates this for robot design optimization, workpiece placement optimization, and a combination of both.
Thus unified the problem can be formulated as a modified trajectory optimization problem:
\begin{equation}
        \begin{aligned}
                & \underset{q(t_0),u}{\text{minimize}} & & \int_{t_0}^{t_f} wL_t(p) + L_d(p,q,u) dt \\
                & \text{subject to} & &  \dot{p} = J(q) \dot{q} \\
                & & & \dot{q}_m = u\\
                & & & \dot{q}_d = 0\\
        \end{aligned}
        \label{eq:to_problem}
\end{equation}
where $q$ is the full joint vector $q=(q_m,q_d)^T$, $q_m$ describes the moving joints and $q_d$ the design joints.
The vector $p$ describes the task space position of the robot end effector and is updated using the Jacobian $J(q)$.
The optimization variable $u$ describes the joint velocities of the moving joints.
The scalar $w$ is a weighting factor that can be used to balance the tracking and design costs.
In practice, the system is also often subject to numerous equality and inequality constraints such as joint limits or collision constraints.
These are not used to force the end effector to follow a given toolpath, instead, the toolpath is encoded in the tracking cost $L_t$.
\begin{equation}
L_t(p) = \|p(t) - s(t)\|^2
\end{equation}
where $s(t)$ is the desired toolpath.
The design costs such as cycle time, robot size, or manipulability are encoded in the design cost $L_d$.
Examples can be found in \cite{whitman}, \cite{previous_work} or (\ref{eq:design_cost}).
To show that this formulation is useful for optimizing both problems, the next section will introduce
an example problem and show how it can be solved using this formulation.


\section{EXAMPLE APPLICATION}\label{sec:example_application}
To investigate the proposed formulation's effectiveness,
We will consider the example of designing a robotic system for milling a figure eight into a planar surface.
The design objective is to optimize surface quality by minimizing magnitude of the static deformation caused by robot compliance multiplied by applied forces during the toolpath $s(t)$:
\begin{equation}
        L_d(q,s)= \|C(q) F(s(t))\|^2
        \label{eq:design_cost}
\end{equation}
A robot's compliance $C$ is dependent on its pose through the Jacobian $J(q)$ of its forward kinematics and joint stiffness matrix $K$~\cite{stiffness_placement}
\begin{equation}
C(q) = J(q)^T K^{-1} J(q).
\end{equation}
All moving joints $q_m$ are assumed to be equally stiff, which is why their corresponding value in $K$ can be simply set to 1 for optimization purposes.
The design joints $q_d$ are assumed to be infinitely stiff.
The force $F(s(t))$ acting on the robot is a result of the tool biting into the material and thus dependent on the toolpath.
Assuming the tool is always fully engaged, the force can be approximated from the force model described in~\cite{milling_force} as follows:
\begin{equation}
F(s(t)) = \alpha \frac{\dot{s}(t)}{\|\dot{s}(t)\|}
\end{equation}
where $\alpha$ is a constant that depends on various process parameters assumed constant such as material and rotational tool speed.

To model the irrelevance of the milling tools z-orientation, we modify the tracking objective to only keep the tool oriented normally to the surface while following the toolpath:
\begin{equation}
        L_t(p) = \|p_t(t) - s_t(t)\|^2+\|p_z(t) - s_z(t)\|^2
\end{equation}
where $p_t$ and $p_z$ are the current position and third collum of the current rotation respectively.
Similarly $s_t$ and $s_z$ are the target position and third collum of the target rotation of the robot.
To optimize the robotic system we also need to define the design space.
In the context of the trajectory optimization formulation, this means defining the design joints and their limits.
The placement of the workpiece can be modeled using three prismatic and three revolute joints.
While these joints move the robot this is mathematically equivalent to moving the workpiece.
The prismatic joints enable movement in the x, y, and z directions, constrained within a cuboid representing the work cell's available space.
Three variations of the design space are considered.
Firstly, the optimal design and placement of a conventional industrial robot with six revolute joints are examined, modeled using Denavit-Hartenberg (DH) parameters.
Since only continuous DH parameters are optimized we will call this problem the continuous design problem.
Secondly, the design space is expanded to explore different kinematic structures, including joint type and number.
This will be called the kinematic structure design problem.
In both problems, the design joints representing physical size (DH parameters $d$ or $a$) are constrained to be positive,
and a regularization term $L_r$ is added based on \cite{whitman} that penalizes large values of $d$ or $a$ to prevent excessively large robots.
Lastly, the modular design problem investigates the design and placement of a modular robot.
Each module $k$ is modeled by a set of DH parameters $q_j$.
This example will use five different modules outlined in Table~\ref{tab:modules}.
An overview of the three design spaces is provided in Table~\ref{tab:designs}.

\begin{table}
\begin{tabular}{l|l}
        Design Problem & Robot Designspace \\
        \hline
        Continuous & $A_c:=\{d_i, a_i, \alpha_i |i\in{1,\ldots,6} \}$\\
        Kinematic Structure & $A_s:=\{n,\{d_i\wedge \theta_i, a_i, \alpha_i |i\in{1,\ldots,n} \}|n\in{1,\ldots,6}\}$\\
        Modular & $A_m:=\{$Modules from Table~\ref{tab:modules}$\}$\\
\end{tabular}
\caption{Design space for each design problem}\label{tab:designs}
\end{table}

\begin{table}[]
\centering
\begin{tabular}{ccccc}
        Module & $\theta$ & d   & a   & $\alpha$ \\ 
        \hline
        \begin{minipage}{.1\textwidth}
        \includegraphics[width=\linewidth]{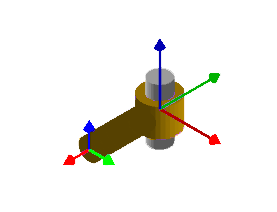}
        \end{minipage}&*                     & 0   & 0.5 & 0                     \\
        \begin{minipage}{.1\textwidth}
        \includegraphics[width=\linewidth]{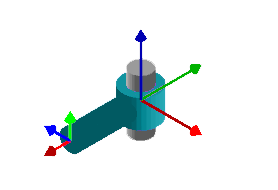}
        \end{minipage}&*                     & 0   & 0.5 & $\pi/2$  \\
        \begin{minipage}{.1\textwidth}
        \includegraphics[width=\linewidth]{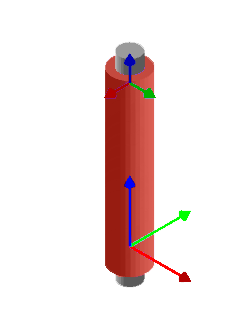}
        \end{minipage}&*                     & 1   & 0   & 0                     \\
        \begin{minipage}{.1\textwidth}
        \includegraphics[width=\linewidth]{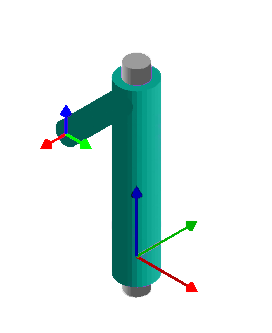}
        \end{minipage}&*                     & 1   & 0.5 & 0                     \\
        \begin{minipage}{.1\textwidth}
        \includegraphics[width=\linewidth]{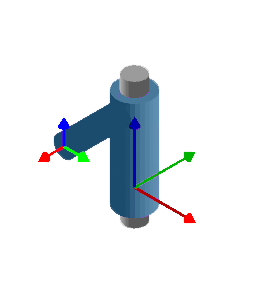}
        \end{minipage}&*                     & 0.5 & 0.5 & $-\pi/2$
\end{tabular}
\caption{Modules in the modular design problem}\label{tab:modules}
\end{table}

\section{Solving the Design Problems}
In this section, we will solve the design problems we have formulated.
We begin with the continuous design problem, which can be directly formulated as a trajectory optimization problem.
Subsequently, we demonstrate how the solution for the continuous problem can be extended to address the kinematic structure design problem and the modular design problem.
\subsection{Solving the continuous design problem}
The continuous design problem takes the form of the trajectory optimization problem defined in (\ref{eq:to_problem}).
To tackle this problem, we employ the direct collocation method~\cite{collocation}.
Direct collocation is a type of trajectory optimization algorithm in which the joint trajectory is approximated using a set of piecewise polynomials.
Instead of directly optimizing the joint trajectory, collocation optimizes the parameters of these polynomials.
This results in a very sparse optimization problem that scales well with increasing degrees of freedom~\cite{collocation}.
By setting the polynomial degree to 3 for moving joints and 0 for design joints,
we ensure a smooth moving joint trajectory while keeping the design joints constant.
This implicitly satisfies the design joint constraint $\dot{q}_d=0$.
All of these reasons make direct collocation particularly well suited for the continuous design problem.
Solving the collocation yields the optimal joint trajectory $q_m$ and optimal design joints $q_d$.

\subsection{Solving the kinematic structure design problem}
Solving the kinematic structure problem involves determining the values of continuous design joints, identifying whether a joint is prismatic or revolute, and selecting the number of joints.
As such it is a direct extension of the continuous design problem that adds a number of discrete decisions.
While this might initially seem like a complex mixed integer problem, for a 6-DOF robot there are only 64 possible kinematic structures.
And due to exponential scaling, considering all robots with 6 or fewer joints only yields 123 such structures.
Given the efficient nature of solving the continuous design problem, it is feasible to simply solve it for each structure.
The kinematic structure design problem can thus be solved by solving the continuous design problem for each possible structure and then selecting the best one.
Best in this case means the structure that yields the lowest design cost while still being able to track the toolpath.
Although this might seem like a very naive approach, it can easily be parallelized and is guaranteed to find the optimal structure given a suitable continuous design problem solver.

\subsection{Solving the modular design problem}\label{sec:modular}
The modular design problem initially appears quite different from the previous two problems because it only has discrete design joints.
And we also can't use the same approach as in the kinematic structure design problem because the number of possible robots is now much larger.
Instead, we propose to just assume that the design joints are continuous, then solve the continuous design problem, and then round the solution to the nearest modular robot.
For this to work, it is crucial that the continuous solutions closely resemble the modules.
This is achieved by introducing an additional design cost, defined as:
\begin{equation}
        L_{mod} = \sum_{i=1}^{6} \prod_{k=1}^{K}  \|q_{d,i} - q_{k}\|^2
        \label{eq:modular}
\end{equation}
Here, $q_{d,i}$ represents the vector describing the design joints in the $i$-th DH transformation, and $q_j$ represents the vector describing the DH parameters of module $j$.
This function is minimized if each set of design joints $q_{d,i}$ is equal to one of the modules $q_j$.
In practive additional weights are added to balance the influence of the angular and linear components of $q_{d,i}$.
Depending on the size of the robot one can weigh the angular and linear components of $q_{d,i}$ differently to prevent the angular components from dominating the cost.
After solving this modified continuous design problem each design joint $q_{d,i}$ is rounded to its closest module $q_j$.
The placement problem is then solved again to account for potential changes resulting from the rounding process.
This process assumes that modules share the same joint type and that the number of moving joints is predetermined.
If this is not the case continuous optimization can be extended  to the kinematic structure design problem.

\section{Results}\label{sec:experiments}
In this paper's introduction, we asserted that the robot design problem is intertwined with the robot placement problem.
In this section, we examine this claim using the example application of an optimal milling system.
We expect that an optimally placed optimized robot would outperform and look different when compared to an optimized robot placed at the origin.
For the optimization results to be meaningful, the performance of the optimizer needs to be benchmarked first.
Therefore, we will conduct a series of experiments to evaluate the performance of the trajectory optimization approach for the continuous design problem.
We will then investigate how the trajectory optimization approach performs on the kinematic structure design problem and the modular design problem.

\subsection{Results for the continuous design problem}
\begin{figure}
        \centering
        \includegraphics[width=\columnwidth]{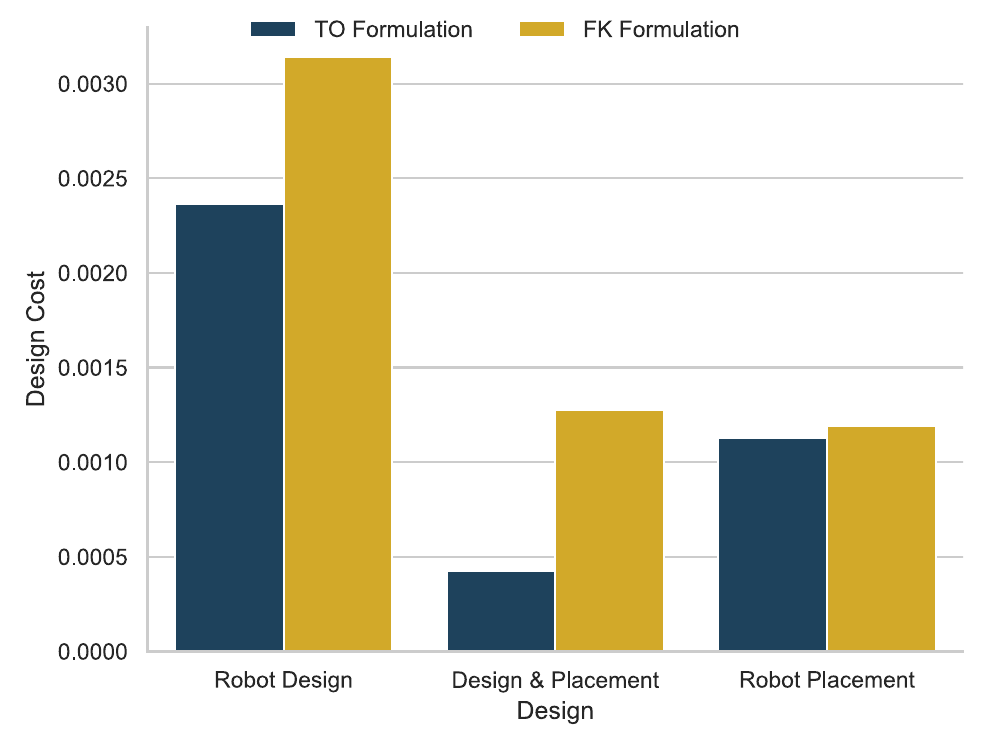}
        \caption{Results of the 6R robot design and workpiece placement problem}\label{fig:results}
\end{figure}

\begin{figure}
        \centering
        \begin{subfigure}[b]{0.4\columnwidth}
                \includegraphics[width=\textwidth]{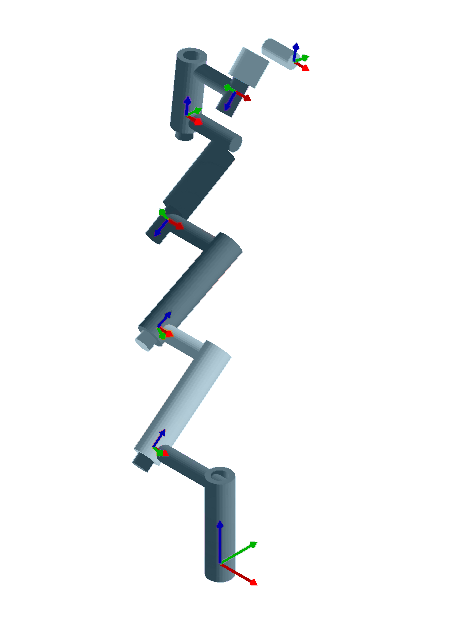}
                \caption{Optimized Robot}\label{fig:subfig1}
              \end{subfigure}
              \hfill
              \begin{subfigure}[b]{0.4\columnwidth}
                \includegraphics[width=\textwidth]{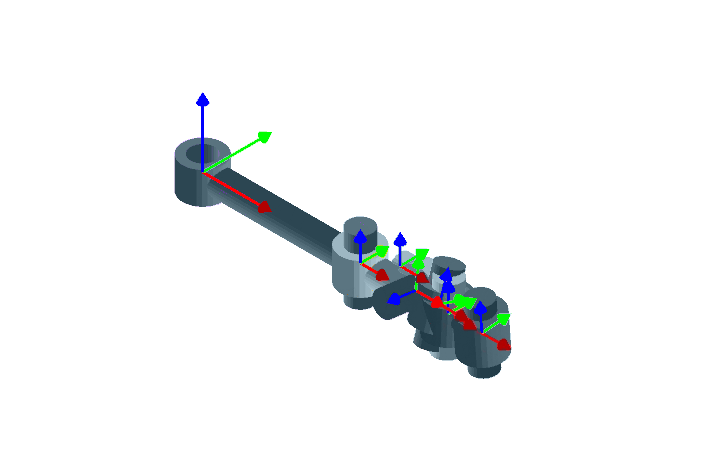}
                \caption{Optimized Robot and Placement}\label{fig:subfig2}
              \end{subfigure}

              \begin{subfigure}[b]{\columnwidth}
                \center
                \includegraphics[height=100pt]{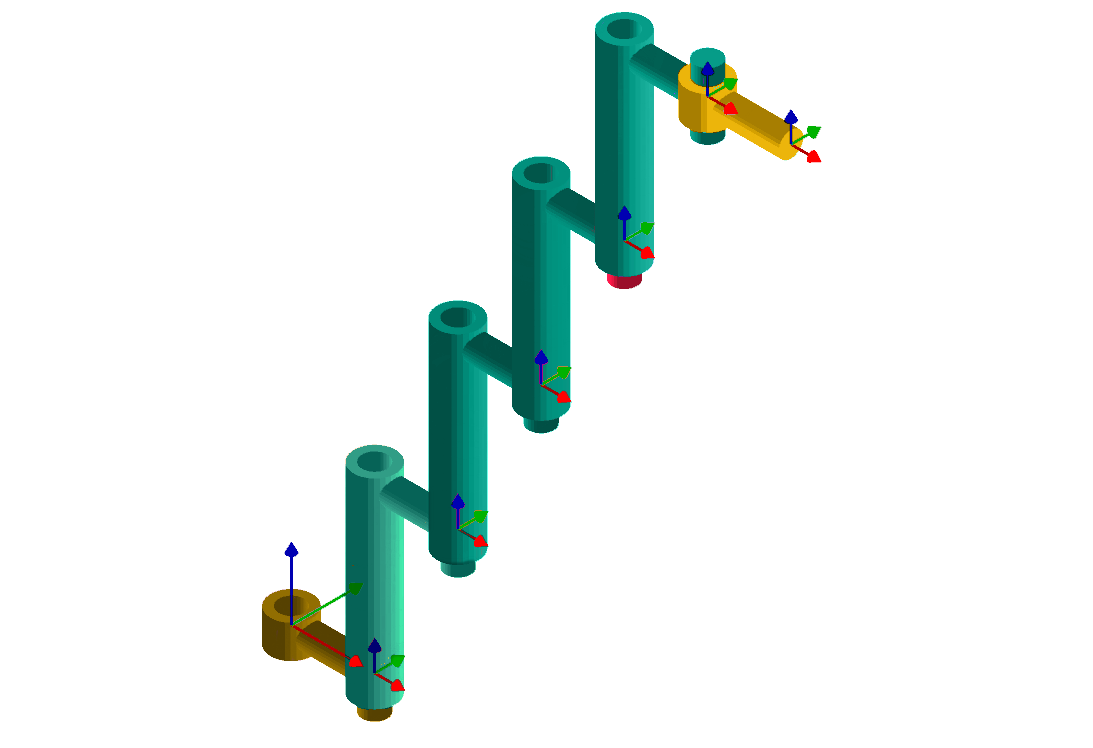}
                \caption{Optimized Modules}\label{fig:subfig3}
              \end{subfigure}
        \caption{Optimized Robots with joint angles $\theta_i=0~\forall~i\in{1,\ldots,6}$}\label{fig:optimal_robot}
\end{figure}

Before investigating how combining placement and design optimization affects performance we first benchmark the trajectory optimization approach (TO) against existing state of the art approaches.
Here we use the forward kinematics based optimization  approach (FK) proposed by Whitman et al.~\cite{whitman}.
In its continuous form it is given by:
\begin{equation}
        \begin{aligned}
                & \underset{q_m,q_d}{\text{minimize}} & & \int_{t_0}^{t_f} L_d(q_m,q_d) dt \\
                & \text{subject to} & &  L_t(p(q_m,q_d)) \leq \epsilon \\
        \end{aligned}
        \label{eq:nlp_problem}
\end{equation}
where $\epsilon$ is a small constant and $L_t$ is the tracking cost defined in (\ref{eq:to_problem}).
While this formulation was proposed for robot design optimization, using the notion of design joints it can also be used for placement optimization.
For this reason, the FK approach will be used as a benchmark for both the placement problem and the combined problem.
The resulting total design cost (\ref{eq:design_cost}) integrated over the toolpath $s$ is shown in Fig.~\ref{fig:results}.
Since both algorithms require a discretization of the toolpath $s$ we use 80 waypoints for both algorithms.
Fig,~\ref{fig:results} shows that the TO approach outperforms the FK approach in each case with the gap increasing as the number of design joints increases.
This indicates that formulating the Design Problem as a trajectory optimization problem is a promising approach.
The Figure also shows that the combined design and placement yields better results no matter the algorithm.
One might simply attribute this to the fact that the combined problem has more degrees of freedom.
However placement alone performs better than the design optimization alone, even though it has much fewer degrees of freedom.
This indicates that placement is an integral part of the design process, in some cases even more important than the design of the robot itself.
Looking at the resulting robots which are shown in Fig.~\ref{fig:optimal_robot} also see that the jointly optimized and placed robot looks very different from the robot optimized at the origin.
This further indicates that the placement problem is not independent of the robot design problem and needs to be considered when optimizing a robot.

\subsection{Results for the kinematic structure design problem}
\begin{figure}
        \centering
        \includegraphics[width=\columnwidth]{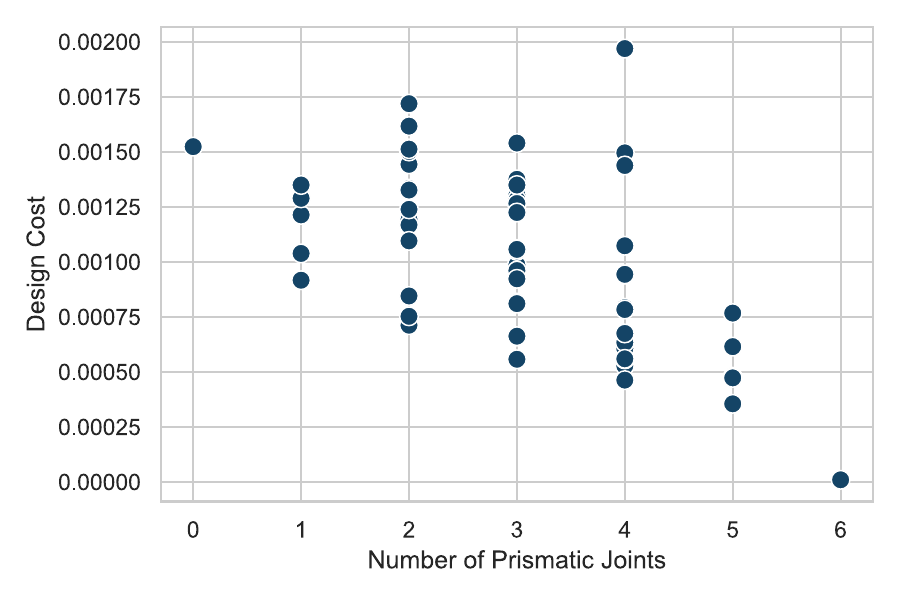}
        \caption{Deformation cost for the optimal 6-DOF robot given the number of prismatic joints}\label{fig:kinematic_structure}
\end{figure}

\begin{figure}
        \centering
        \includegraphics[width=\columnwidth]{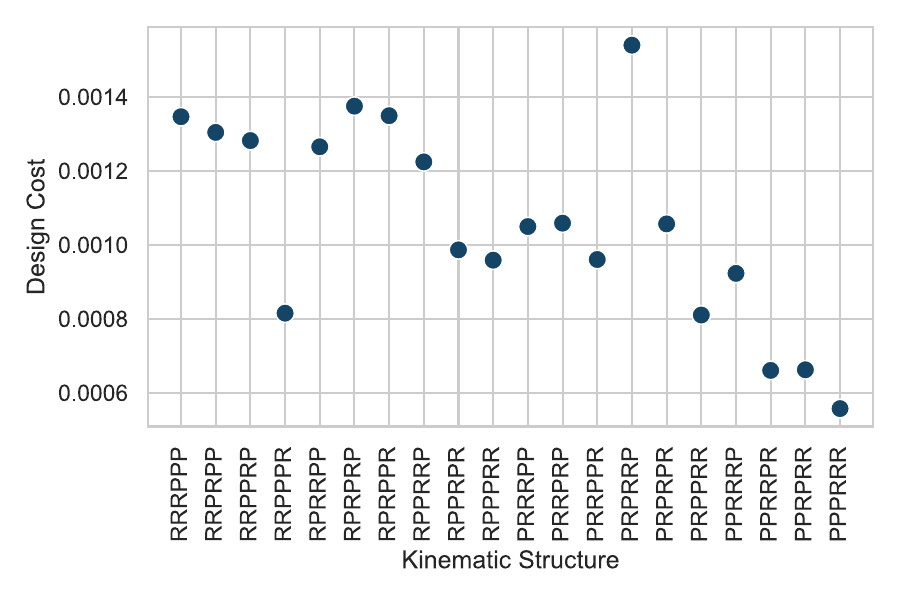}
        \caption{Deformation cost for the optimal 6-DOF robot with three prismatic joints}\label{fig:kinematic_structure_3}
\end{figure}

\begin{figure}
        \centering
        \includegraphics[width=\columnwidth]{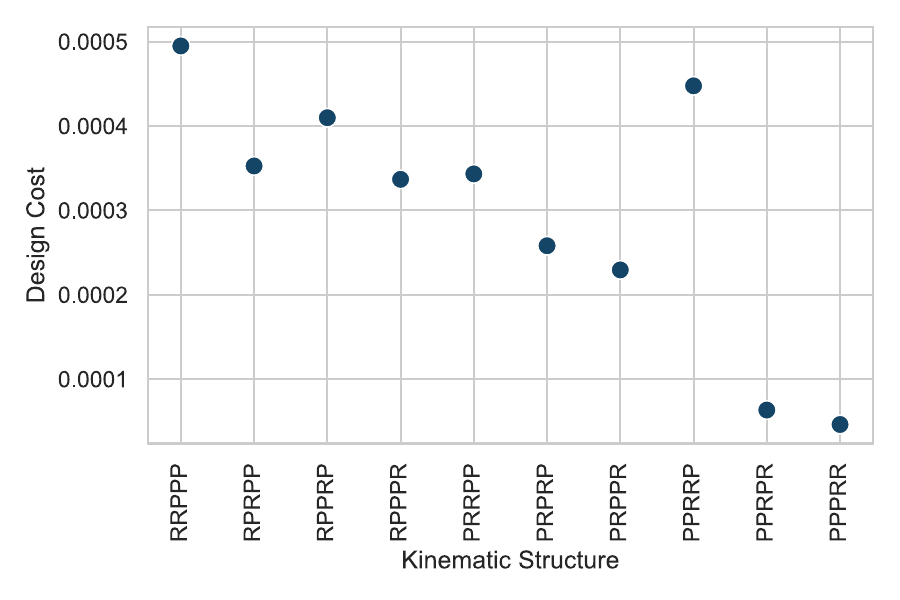}
        \caption{Deformation cost for the optimal 5-DOF robot with three prismatic joints}\label{fig:kinematic_structure_5_3}
\end{figure}

To assess the performance of the proposed solution to the kinematic structure design problem, we will compare it to the expected optimal solution.
In this case we assume that for surface milling the optimal structure is given by serial milling machines which employ prismatic joints near the base for positioning and revolute joints solely for tool reorientation.
This is done to minimize the leverage acting on revolute joints causing them to be more prone to deformation.
To evaluate the optimization's agreement with the expected solution,
we optimize each of the 64 possible 6-degree-of-freedom (6-DOF) kinematic structures for our milling task.
As the milling system operates on a planar surface, no revolute joints are necessary for tool reorientation.
We thus anticipate a preference for structures with fewer revolute joints.
Fig.~\ref{fig:kinematic_structure} thus shows the design cost plotted against the number of prismatic joints in the structure which confirms our expectations.
To check whether our optimizer finds the optimal position of the revolute joints near the end effector we analyze the positioning of revolute joints in the 3-revolute-3-prismatic (3R3P) structure (Fig.~\ref{fig:kinematic_structure_3}).
We can see that the optimizer correctly finds the optimal position of the revolute joints near the end effector.
We repeat the optimization process for structures with fewer joints, observing a consistent trend as seen in Fig.~\ref{fig:kinematic_structure_5_3} for the 5-degree-of-freedom (5-DOF) example.
As expected the 5-DOF robot outperformed the 6-DOF robot due to the latters additional joint's potential for introducing deformation.
This trend continues for the 4-DOF and 3-DOF cases, with the 3-DOF PPP structure being the optimal solution for our milling task.
Consequently precisely the structure of simple face milling machines.
In conclusion, the structure optimizer correctly identifies the optimal kinematic structure for our milling task.
Moreover, optimizing the structure leads to significant performance improvement compared to continuous design optimization.

\subsection{Results for the modular design problem}
\begin{figure}
        \centering
        \includegraphics[width=0.5\textwidth]{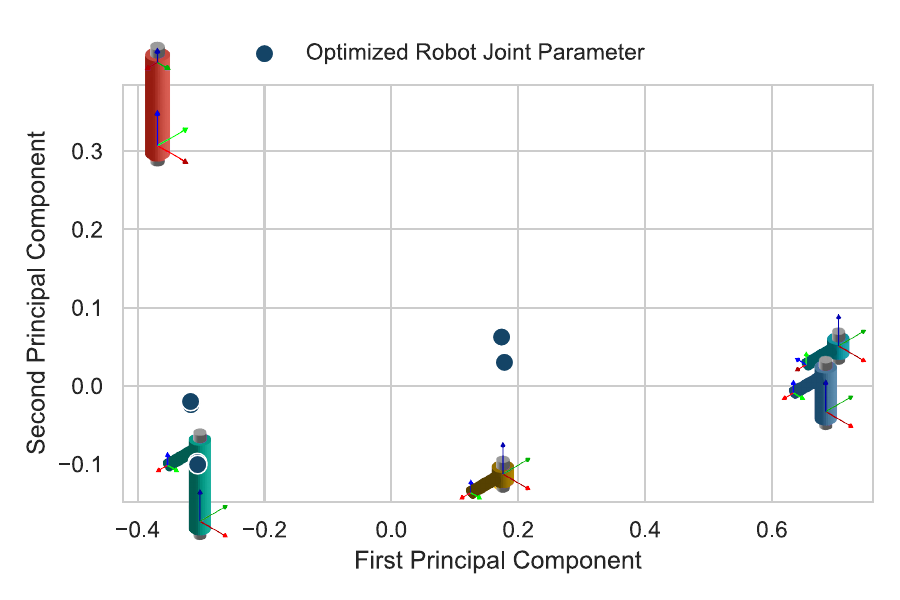}
        \caption{PCA plot of the design joints $q_{d,i}$ and the modules $q_j$.}\label{fig:module_tsne}
\end{figure}

While existing approaches like~\cite{whitman_rl} offer modular robot design methods, they cannot be extended to incorporate placement, as mentioned in the introduction.
Hence, we can't use them as benchmarks for our modular design approach.
However, our TO approach has already demonstrated its capability to optimize both robot design and placement.
We thus mainly have to check whether introducing modular design cost (\ref{eq:modular}) results in solutions that resemble the modules and how this affects the overall performance.
To address this question we perform the modular design optimization and plot the resulting sets of design joints $q_{d,i}$ and modules $q_j$ in a PCA plot (Fig.~\ref{fig:module_tsne}) to access their similarity.
As we can see most joints can be clearly assigned to a module, indicating that introducing the modular design cost (\ref{eq:modular}) results in solutions that resemble the modules.
Note that the resulting modular robot looks very different from the original TO formulation as pictured in Fig.~\ref{fig:optimal_robot}.
Its performance however is similar with a design cost $\num{6.72e-4}$ compared to the original $\num{4.16e-4}$. It also still outperforms the FK method with $\num{1.261e-3}$.
This indicates that our modular design approach is capable of finding suitable solutions and can extend modular robot design to incorporate placement optimization.

\section{Limitations and future prospects}\label{sec:limitations}
The experiments presented in the previous section have shown that optimizing the design of a robot should also consider its placement.
However, we believe that the field of robot and environment design optimization is still in its infancy and there are many open questions.
In this section, we will outline some of these, highlighting the limitations of our approaches and offer future research directions.
One area requiring further investigation is the integration of design joints into trajectory planning.
The number and gaps between waypoints seem to greatly influence performance inversely to expectations, but the reasons remain unclear.
Additionally, planning complex paths between distant waypoints, especially in the presence of intricate obstacles,
presents challenges that collision costs like the one proposed in~\cite{whitman}
may not adequately address.
Integrating design joints into gradient-free optimizers like RRT is an open question.
In modular design, the influence of available modules on optimization performance and the impact of the number of modules are important considerations yet to be explored.
A general practical concern is that current algorithms provide only a single solution, while different scenarios may have multiple equally valid solutions.
To provide a comprehensive picture, presenting a set of solutions would be valuable.

\section{CONCLUSION}
In this work, we have taken a look at the long-standing problems of task-specific robot design and workpiece placement.
We highlighted that both problems can be thought of as trajectory optimization, allowing for their unification.
Using the example of a milling system we showed that combining these problems leads to a significant performance improvement.
However, the collocation algorithm used to achieve this is only one of many possible TO approaches.
Using the TO formulation many different algorithms can be applied.
But no matter the algorithms used, we have hopefully made the case that the design of the robot and the design of its environment should no longer be treated as separate problems.
Because leveraging the synergy between these two problems can create much more capable robotic systems.

\bibliographystyle{IEEEtran}
\bibliography{IEEEabrv,bibliography}

\end{document}